\documentclass{esannV2}
\usepackage[latin1]{inputenc}
\usepackage{amssymb,amsmath,array}

\usepackage{graphicx}
\usepackage{pgfplots}
\usepackage{tikz}
\usetikzlibrary{calc}
\usetikzlibrary{positioning}

\usepackage{subcaption}

\usepackage{multirow}

\begin{document}
\title{Real-time convolutional networks for sonar image classification in low-power embedded systems}

\author{Matias Valdenegro-Toro
%
%
\vspace{.3cm}\\
%
Ocean Systems Laboratory - School of Engineering \& Physical Sciences\\
Heriot-Watt University, EH14 4AS, Edinburgh, UK\\
Email: m.valdenegro@hw.ac.uk \\
}

\maketitle

\begin{abstract}
Deep Neural Networks have impressive classification performance, but this comes at the expense of significant computational resources at inference time. Autonomous Underwater Vehicles use low-power embedded systems for sonar image perception, and cannot execute large neural networks in real-time. We propose the use of max-pooling aggressively, and we demonstrate it with a Fire-based module and a new Tiny module that includes max-pooling in each module. By stacking them we build networks that achieve the same accuracy as bigger ones, while reducing the number of parameters and considerably increasing computational performance. Our networks can classify a $96 \times 96$ sonar image with $98.8 - 99.7 \%$ accuracy on only 41 to 61 milliseconds on a Raspberry Pi 2, which corresponds to speedups of $28.6 - 19.7$.

\end{abstract}

\section{Introduction}

Convolutional Neural Networks (CNNs) have revolutionized object detection and recognition \cite{krizhevsky2012imagenet}, enabling classifiers that generalize very well outside of their training sets, and recent advances allow easy representation of complex functions through increasing number of layers in a model \cite{glorot2011deep}.

\begin{figure*}[!htb]
    \centering
    \begin{subfigure}{0.15 \textwidth}
        \includegraphics[width = \textwidth]{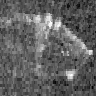}
        \caption{Can\vspace*{0.4cm}}
    \end{subfigure}
    \begin{subfigure}{0.15 \textwidth}
        \includegraphics[width = \textwidth]{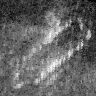}
        \caption{Plastic\\Bottle}
    \end{subfigure}
    \begin{subfigure}{0.15 \textwidth}
        \includegraphics[width = \textwidth]{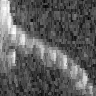}
        \caption{Chain\vspace*{0.4cm}}
    \end{subfigure}
    \begin{subfigure}{0.15 \textwidth}
        \includegraphics[width = \textwidth]{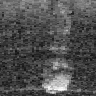}
        \caption{Drink\\Carton}
    \end{subfigure}
    \begin{subfigure}{0.15 \textwidth}
        \includegraphics[width = \textwidth]{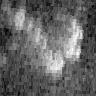}
        \caption{Hook\vspace*{0.4cm}}
    \end{subfigure}

    \begin{subfigure}{0.15 \textwidth}
        \includegraphics[width = \textwidth]{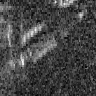}
        \caption{Glass\\Bottle}
    \end{subfigure}
    \begin{subfigure}{0.15 \textwidth}
        \includegraphics[width = \textwidth]{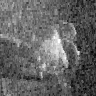}
        \caption{Propeller\vspace*{0.4cm}}
    \end{subfigure}
    \begin{subfigure}{0.15 \textwidth}
        \includegraphics[width = \textwidth]{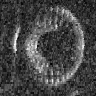}
        \caption{Tire\vspace*{0.4cm}}
    \end{subfigure}
    \begin{subfigure}{0.15 \textwidth}
        \includegraphics[width = \textwidth]{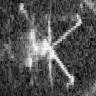}
        \caption{Valve\vspace*{0.4cm}}
    \end{subfigure}
    \caption{Sample Forward-Looking Sonar Images from our train/test sets.}
    \label{sampleFLSImages}
\end{figure*}

Neural network complexity increases with depth, but so does computation times. Graphics Processing Units (GPUs) are used to offset such increases, and can achieve real-time computational performance at inference time. CNNs are ideal for detection \cite{Valdenegro-Toro2016} and recognition tasks performed by Autonomous Underwater Vehicles (AUV) on sonar images (Fig. \ref{sampleFLSImages}).

A key limitation is power, as AUVs use batteries for long-term autonomy, and good power management is important in order not to constraint autonomy and mission time. Low-power embedded systems are preferred, and the power budget does not allow the use of GPUs. Low cost is also an issue as GPUs require considerable heat management. The motivation for this work is to enable the execution of a classification CNN (inference time) on sonar data captured in real-time (up to 15 Hz) on a low power embedded device.

A large literature exists about model compression \cite{he2015convolutional} \cite{han2015learning} 
\cite{han2015deep} \cite{denton2014exploiting}, where a machine learning model is compressed in order to reduce the number of parameters, typically by slightly reducing model performance. A reduction of model parameters usually translates in a decrease of computation time, but this relationship is complex and non-linear.

Denton et al. \cite{denton2014exploiting} uses linear structure in learned convolutional filters to approximate them, obtaining speedups of up to 2.0. He et al. \cite{he2015convolutional} manually designs CNN architectures in order to reduce the number of required operations while maintaining network complexity. Han et al. \cite{han2015learning} learns only important connections and weights and prunes the rest, reducing parameter count by an order of magnitude, and obtains speedups up to 5.0. Han et al. \cite{han2015deep} prunes and quantizes weights, and then compressed the model with Huffman coding, reducing the number of parameters by 50, and speedups in the order of 3.0. Most of these techniques concentrate in reducing the number of parameters and the model file size, for mobile and storage constrained applications. Our approach is different as we only need to reduce computation time to under 66 ms, as storage is not a problem for AUVs.

In this paper we propose two new neural network modules that can be used to build an image classification architecture with a low parameter count, which translates into large computational performance improvements without considerably decreasing accuracy. We evaluate this architecture in our sonar image dataset (as shown in Fig. \ref{sampleFLSImages}). Our contributions are the two new Tiny and SmallFire modules, and neural network architectures that can classify $96 \times 96$ sonar images with high accuracy in real-time on a Raspberry Pi 2.

\section{Small Network Architectures for Sonar Image Classification}

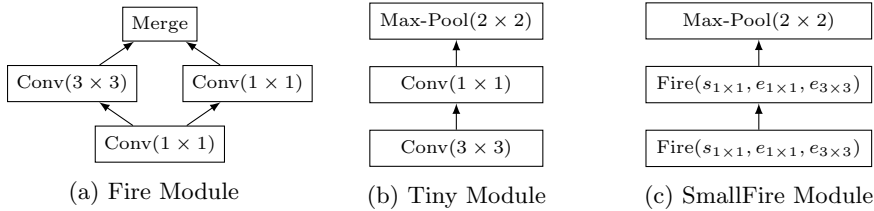
\begin{figure}        
    \centering
    \begin{subfigure}{0.32 \textwidth}
        \centering
        \begin{tikzpicture}[style={align=center, minimum height=0.4cm}]
        \node[] (dummy) {};
        \node[draw, below=1em of dummy](B) {{\scriptsize Conv($1 \times 1$)}};
        \node[draw, left=0.5em of dummy] (C) {{\scriptsize Conv($3 \times 3$)}};
        \node[draw, right=0.5em of dummy] (D) {{\scriptsize Conv($1 \times 1$)}};
        \node[draw, above=1em of dummy](CONC) {{\scriptsize Merge}};
        
        \draw[-latex] (B) -- (C);
        \draw[-latex] (C) -- (CONC);
        \draw[-latex] (B) -- (D);
        \draw[-latex] (D) -- (CONC);
        \end{tikzpicture}
        \caption{Fire Module}
    \end{subfigure} 
    \begin{subfigure}{0.32 \textwidth}
        \centering
        \begin{tikzpicture}[style={align=center, minimum height=0.4cm, minimum width = 2.3cm}]
        \node[draw] (A) {{\scriptsize Conv($3 \times 3$)}};
        \node[draw, above=1em of A] (B) {{\scriptsize Conv($1 \times 1$)}};
        \node[draw, above=1em of B] (C) {{\scriptsize Max-Pool($2 \times 2$)}};
        \draw[-latex] (A) -- (B);
        \draw[-latex] (B) -- (C);
        \end{tikzpicture}
        \caption{Tiny Module}        
    \end{subfigure}
    \begin{subfigure}{0.32 \textwidth}
        \centering
        \begin{tikzpicture}[style={align=center, minimum height=0.4cm, minimum width = 3.0cm}]
        \node[draw] (A) {{\scriptsize Fire($s_{1 \times 1}, e_{1 \times 1}, e_{3 \times 3}$)}};
        \node[draw, above=1em of A] (B) {{\scriptsize Fire($s_{1 \times 1}, e_{1 \times 1}, e_{3 \times 3}$)}};
        \node[draw, above=1em of B] (C) {{\scriptsize Max-Pool($2 \times 2$)}};
        \draw[-latex] (A) -- (B);
        \draw[-latex] (B) -- (C);
        \end{tikzpicture}
        \caption{SmallFire Module}        
    \end{subfigure}
    \caption{Fire module \cite{iandola2016squeezenet} and our proposed Tiny and SmallFire module}
    \label{moduleArchitectures}
\end{figure}

Our architecture is based on the Fire module from SqueezeNet \cite{iandola2016squeezenet}. The Fire module uses $1 \times 1$ filters for a squeeze stage, followed by an expand stage that uses $1 \times 1$ and $3 \times 3$ filters, as shown in Fig. \ref{moduleArchitectures}a. The squeeze stage has $s_{1 \times 1}$ filters of $1 \times 1$ spatial size, while the expand stage has $e_{1 \times 1}$ and $e_{3 \times 3}$ filters of the corresponding spatial sizes.

Our proposed module, denominated the Tiny module, contains an equal number of $1 \times 1$ and $3 \times 3$ filters, starting with $3 \times 3$ to capture spatial relations in the input data, and $1 \times 1$ filters for a small parameter count increase without capturing spatial relations. Alternatively a smaller number of $1 \times 1$ filters could be used for dimensionality reduction, but this option greatly reduced accuracy in our experiments. Connecting the $3 \times 3$ filters to the input (in contrast to the initial $1 \times 1$ filters in the Fire module) allows to skip an expensive initial convolution that is required in Fire-based networks.
 
Finally, as part of the module itself, the output of the $1 \times 1$ convolution is max-pooled with $2 \times 2$ non-overlapping cells. This aggressive down-sampling of feature maps allows for performance increases, as subsequent layers have to process less data.

A Batch normalization \cite{ioffe2015batch} layer is added between the final $1 \times 1$ convolution and the max-pooling layers for regularization and training acceleration.

To instantiate a CNN architecture with the Tiny module, $n$ modules are stacked, with the initial module connected to the input data. After the specified number of modules, a single $1 \times 1$ convolution with an $c$ filters is performed, where $c$ is equal to the number of classes . Global average pooling is then applied, which reduces the $c$ feature maps to a vector of $c$ elements. Then the softmax activation is applied to produce the network output.

The network then can be trained from randomly initialized weights, with a cross-entropy loss function. We use the ADAM optimizer \cite{kingma2014adam} with a learning rate $\alpha = 0.1$, for 30 epochs, with a batch size of 128 elements.

\section{Experimental Evaluation}

We evaluated our network architecture in dataset of 334900 Forward-Looking Sonar images, with 11 classes in total (including a background class). Each image has size $96 \times 96$ pixels. We performed 5-fold Cross Validation to evaluate robustness and compare different networks. Accuracy is reported as mean and standard deviation over folds. Computational evaluation of all networks at inference time was performed in a Raspberry Pi 2, using Keras with the Theano backend. The standard deviation of all time measurements was under 3 milliseconds. Only a single core was used in order to save power.

Our baseline is a CNN based on LeNet \cite{lecun1998gradient}, with configuration Conv(32, 5, 5)-MaxPool(2, 2)-Conv(32, 5, 5)-MaxPool(2, 2)-FC(64)-FC(11). This network has 930K parameters and obtains $98.8 \pm 0.4 $\% test accuracy, with a computation time of 1200 ms per image.

We also designed two Fire-based architectures, one as a baseline, and another to evaluate the limits of the Fire module with a low number of convolution filters, which we named SmallFireNet. The baseline Fire architecture has two Fire modules with $s_{1 \times 1} = 16, e_{1 \times 1} = 16, e_{3 \times 3} = 16$, an initial $5 \times 5$ convolution with 8 filters, and an output $5 \times 5$ convolution with $c$ filters. Global average pooling and softmax are used to produce the final output. This baseline architecture has 18K parameters and obtains $99.6 \%$ accuracy, which is the highest we have obtained in our dataset. Computation time for this baseline is 600 ms per image.

The SmallFireNet architecture has a variable number of SmallFire modules (shown in Fig. \ref{moduleArchitectures}) with $s_{1 \times 1} = e_{1 \times 1} = e_{3 \times 3} = 4$, with the same input and output convolutions configurations as the Fire baseline. We evaluated the use of $n \in [1, 2, 3]$ SmallFire modules in this architecture. One big difference between the Fire module in \cite{iandola2016squeezenet} and our SmallFire module is that we use $2 \times 2$ Max-Pooling after every two Fire modules, in concordance with our design strategy of aggressive down-sampling

We evaluated several instances of our Tiny module, in a network we named TinyNet. 4 or 8 convolution filters with $n \in [1, 2, 3, 4, 5]$ number of Tiny modules in the network. This allows us to evaluate the trade-off between number of fire modules with computational performance and classification accuracy.

\begin{table}
	\centering
	\begin{tabular}{|l|c|c|c|c|c|c|}
		\hline
		        & \multicolumn{3}{|c|}{TinyNet - 4 Filters} & \multicolumn{3}{|c|}{TinyNet - 8 Filters}\\
		\hline
		\#   	& Params & Accuracy  	     & Time  & Params & Accuracy & Time  \\
		\hline 
		1 		& 307 	 & $93.5 \pm 0.4$ \% & 28 ms & 443  & $95.8 \pm 0.5$ \% & 57 ms\\ 
		\hline 
		2  		& 571    & $95.0 \pm 0.7$ \% & 35 ms & 1195 & $98.2 \pm 0.3$ \% & 88 ms\\ 
		\hline 
		3 		& 787    & $95.9 \pm 0.3$ \% & 38 ms & 1899 & $98.4 \pm 0.2$ \% & 95 ms\\
		\hline
		4 		& 979	 & $97.0 \pm 0.3$ \% & 40 ms & 2579 & $98.8 \pm 0.2$ \% & 99 ms\\
		\hline
		5 		& 1159   & $98.8 \pm 0.2$ \% & 42 ms & 3247 & $99.6 \pm 0.1$ \% & 110 ms\\
		\hline 
	\end{tabular}
	\caption{TinyNet performance as function of number of modules (\#) and convolution filters (4 or 8). We report mean and standard deviation of accuracy.}
	\label{tinyNet4vsNumberOfModules}. 
\end{table}

\begin{table}
    \centering
    \begin{tabular}{|l|c|c|c|c|c|c|c|c|c|}
    	\hline
    	\# of Modules	& Params & Accuracy		     & Time \\
    	\hline 
    	1 Module		& 3163 	 & $99.0 \pm 0.2$ \% & 70 ms \\ 
    	\hline 
    	2 Modules 		& 3643   & $99.7 \pm 0.2$ \% & 59 ms \\ 
    	\hline 
    	3 Modules		& 4087   & $99.8 \pm 0.1$ \% & 61 ms \\
    	\hline
    \end{tabular}
    \caption{SmallFireNet Performance as function of number of modules, with $e_{1 \times 1} = e_{3 \times 3} = s_{1 \times 1} = 4$. Mean and standard deviation of accuracy is reported.}
    \label{smallFireVsNumberOfModules}
\end{table}

Results as we vary the number of modules are presented in Tables \ref{tinyNet4vsNumberOfModules} and \ref{smallFireVsNumberOfModules}. For TinyNet, four or five modules with four convolutional filters provide the best classification accuracy with a greatly reduced computation time, when compared to our baseline CNN, with a speedup of $28.6$. TinyNet with two modules and eight convolutional filters is also very competitive, with a speedup of $13.6$.

SmallFireNet (Table \ref{smallFireVsNumberOfModules}) also is very competitive, with the highest accuracies in this dataset, and a slightly larger computation time. It is interesting that increasing the number of modules has the effect of decreasing computation time, instead of increasing it as expected. Global average pooling might be driving this effect as adding more modules decreases the size of the feature maps that are input to global average pooling.

\begin{table}
    \centering
    \begin{tabular}{|c|c|c|c|c|}
        \hline 
        Network 		& \# of Params 	& Accuracy 	& Computation Time & Speedup \\ 
        \hline 
        Baseline CNN 	& 930K 				& $98.8$ \% & 1200 ms 		   & N/A \\ 
        \hline 
        Baseline Fire	& 18K 				& \textbf{99.6 \%} & 600 ms 	   & $2.0$ \\ 
        \hline 
        SmallFireNet-3	& 4087 				& \textbf{99.7 \%} & 61 ms 	   & $19.7$ \\ 
        \hline 
        TinyNet-4		& \textbf{1159}		& $98.8$ \% & \textbf{42 ms}   & $\textbf{28.6}$\\ 
        \hline 
        TinyNet-8		& 3247 				& $99.6$ \% & 110 ms		   & $10.9$ \\ 
        \hline 
    \end{tabular} 
    \caption{Summary of our experimental results, evaluated on a Raspberry Pi 2. TinyNet clearly has the biggest speedups, with a small decrease in accuracy. Mean accuracies are reported in this table.}
    \label{resultsSummary}
\end{table}

Table \ref{resultsSummary} shows a summary of our results, including the baselines. TinyNet with 4 convolutional filters is the fastest network, which can be run at almost 24 Hz, and this speedup can be achieved by only sacrificing $0.5 \%$ accuracy. Our results show that by just designing an appropriate CNN architecture, without using fully connected layers, GPUs or multiple cores, one can achieve real-time computational performance on a low power embedded system with small decreases on accuracy. These kind of networks are ideal for use in robot perception.

In the case that a high accuracy is needed ($> 99.0 \%$), the SmallFireNet network is a good choice, as it has very competitive computation time, only $31 \%$ slower than TinyNet-4 and still under our real-time constraint, and the biggest accuracy in our dataset.

Our results also show that max-pooling can be used to greatly decrease the number of parameters in the network, with small decrease in accuracy. TinyNet-4 reduces the number of parameters by a factor of 800, while SmallFireNet reduces it by 227. Both reductions lead to a significant decrease in computation time. It is now clear that there is severe redundancy in the baseline CNN. While varying the number of layers or the filter count in the baseline reduces accuracy, the different architectures in TinyNet and SmallFireNet can exploit the redundancy and successfully perform the task without underfitting.


\section{Conclusions and Future Work}

In this paper we have presented CNN architectures that with little number of layers and convolutional filters, can successfully classify $96 \times 96$ sonar images, with small or no accuracy loss. The small parameter count of these networks allow for a real-time implementation in a low-power embedded system, which is ideal for our specific application of real-time image classification in an AUV.

The best accuracy vs computation time trade-off is given by TinyNet-4, with 98.8\% accuracy and a 28.6 speedup over the naive CNN baseline, while a SmallFireNet-3 has the best accuracy on our dataset at 99.7\%, with a speedup of 19.7. Both networks run at more than 15 Hz in a Raspberry Pi 2, using only a single core.

Our results shows that an alternate approach for model compression is just to train a carefully designed network that uses max-pooling aggresively, and only requires a considerably reduced number of parameters. This is in contrast with results such as SqueezeNet \cite{iandola2016squeezenet}, on which our work is based. Max-pooling is a good tool for parameter reduction and to increase computational performance.

As future work, a clear research direction is to apply other model compression techniques (such as \cite{han2015deep}) for additional parameter reduction and computational speed increases. We also plan to extend this work to regression networks, such as object detection in sonar images.

\begin{footnotesize}


\bibliographystyle{unsrt}
\bibliography{biblio}

\end{footnotesize}


\end{document}